\documentclass{article}

\usepackage{microtype}
\usepackage{graphicx}
\usepackage[normalem]{ulem}
\usepackage{subcaption}
\usepackage{booktabs} % for professional tables
\usepackage{hyperref}
\usepackage{amsmath,amsfonts,bm}

\def\eqref#1{equation~\ref{#1}}
\def\1{\bm{1}}

\DeclareMathAlphabet{\mathsfit}{\encodingdefault}{\sfdefault}{m}{sl}
\SetMathAlphabet{\mathsfit}{bold}{\encodingdefault}{\sfdefault}{bx}{n}

\newcommand{\R}{\mathbb{R}}

\usepackage{xspace}
\usepackage{adjustbox}

\usepackage[preprint]{icml2026}

\usepackage{amsmath}
\usepackage{amssymb}
\usepackage{mathtools}
\usepackage{amsthm}
\newcommand{\thiswork}{ReNCE\xspace}

\usepackage[capitalize,noabbrev]{cleveref}

\theoremstyle{plain}

\theoremstyle{definition}

\theoremstyle{remark}

\usepackage[textsize=tiny]{todonotes}

\icmltitlerunning{ReNCE: Learning to Reason by Noise Contrastive Estimation}

\begin{document}

\twocolumn[
  %\icmltitle{ReNCE: Learning to Reason via Noise Contrastive Estimation}
  \icmltitle{ReNCE: Learning to Reason by Noise Contrastive Estimation}



  \icmlsetsymbol{equal}{*}
  \icmlsetsymbol{current}{$\dagger$}

   \begin{icmlauthorlist}
    \icmlauthor{Wenzheng Zhang}{rutgers}
    \icmlauthor{Karl Stratos}{rutgers,current}
  \end{icmlauthorlist}

  \icmlaffiliation{rutgers}{Department of Computer Science, Rutgers University, New Brunswick, US}

  \icmlcorrespondingauthor{Wenzheng Zhang}{wenzheng.zhang@rutgers.edu}

  \icmlkeywords{RL, Reasoning, RLVR}

  % \begin{icmlauthorlist}
  %   \icmlauthor{Firstname1 Lastname1}{equal,yyy}
  %   \icmlauthor{Firstname2 Lastname2}{equal,yyy,comp}
  %   \icmlauthor{Firstname3 Lastname3}{comp}
  %   \icmlauthor{Firstname4 Lastname4}{sch}
  %   \icmlauthor{Firstname5 Lastname5}{yyy}
  %   \icmlauthor{Firstname6 Lastname6}{sch,yyy,comp}
  %   \icmlauthor{Firstname7 Lastname7}{comp}
  %   %\icmlauthor{}{sch}
  %   \icmlauthor{Firstname8 Lastname8}{sch}
  %   \icmlauthor{Firstname8 Lastname8}{yyy,comp}
  %   %\icmlauthor{}{sch}
  %   %\icmlauthor{}{sch}
  % \end{icmlauthorlist}




  \vskip 0.3in
]



\printAffiliationsAndNotice{$\dagger$ Currently at Apple.}  % no special notice (required even if empty)

\begin{abstract}
  GRPO is a standard approach to endowing pretrained LLMs with reasoning capabilities.
  It estimates the advantage of an outcome from a group of $K$ outcomes, and promotes those with positive advantages inside a trust region.
  Since GRPO discriminates between good and bad outcomes softly,
  it benefits from additional refinements such as asymmetric clipping and zero-variance data filtering.
 While effective, these refinements require significant empirical insight and can be challenging to identify.
 We instead propose an explicit contrastive learning approach.
  Instead of estimating advantages, we bifurcate $K$ outcomes into positive and negative sets, then maximize the likelihood of positive outcomes.
  Our approach can be viewed as an online instantiation of (multi-label) noise contrastive estimation for LLM reasoning.
  We validate our method by demonstrating competitive performance on a suite of challenging math benchmarks against strong baselines such as DAPO and online DPO.
\end{abstract}

\section{Introduction}

GRPO \citep{shao2024deepseekmath} is a standard approach to endowing pretrained LLMs with reasoning capabilities.
It eliminates the need for a separate value network in PPO \citep{schulman2017proximal}
by generating a group of $K$ outcomes and estimating the advantage of each as the within-group standardized reward.
It then promotes good outcomes by maximizing the estimated advantages, while constraining updates to remain close to the previous policy (i.e., inside a trust region)
by ratio clipping and a KL penalty to a reference policy.

GRPO discriminates between good and bad outcomes ``softly'' through advantage-weighted updates.
In practice, additional refinements to the loss, sampling scheme, or reward are made to make these updates more effective.
For instance, DAPO \citep{yu2025dapo} makes ratio clipping asymmetric, filters prompts with zero reward variance, normalizes the loss at a token level, and penalizes overlong responses.
GSPO \citep{zheng2025group} replaces token-level importance ratios with a sequence-level ratio and applies sequence-level clipping to reduce importance-weight variance.
While effective, these refinements require significant empirical insight and can be challenging to identify.

We hypothesize that an explicit contrastive learning approach can be as effective as GRPO.
Instead of estimating advantages, we bifurcate $K$ outcomes into positive and negative sets, then maximize the likelihood of positive outcomes.
Our approach can be viewed as an online instantiation of (multi-label) noise contrastive estimation \citep{gutmann2012noise,mnih2012fast}, henceforth NCE.
We approximate the intractable calculation of the underlying conditional distribution over outcomes
by subsampling negative labels from the model itself---a practice known as ``hard negative mining''.

We call our method ReNCE (Learning to \underline{Re}ason by \underline{NCE}).
In each iteration of ReNCE, we dynamically select prompts whose rollouts are at least 50\% incorrect to optimize for their usefulness.
We then minimize the usual multi-label NCE loss assuming a prompt-rollout score function $s_\theta(x, y)$, with a KL penalty to the previous policy $\pi_{\mathrm{old}}$ to establish a trust region.
Inspired by DPO \citep{rafailov2023direct}, we use the log-ratio score $s_\theta(x,y) \propto \log(\pi_\theta(y|x)/\pi_{\mathrm{old}}(y|x))$ shifted by a reward-based margin to naturally normalize the sequence length by the previous policy.

We empirically validate ReNCE on the task of mathematical reasoning against strong baselines: GRPO, DAPO, and semi-online DPO.
All methods start from the same pretrained LLM checkpoint and train on a combination of the DAPO and MATH datasets \citep{hendrycks2measuring}.
We evaluate their performance on six widely used math benchmarks while ensuring that all baseline results are sufficiently strong to have meaningful comparisons (e.g., $>90$ on Math500).
We find that ReNCE is highly competitive, achieving the highest average performance across the six benchmarks.\footnote{We make our code available at: \url{https://github.com/WenzhengZhang/ReNCE}}
We also perform ablation studies to analyze the importance of our design choices.

\section{Related Work}

We discuss related work to better contextualize our contributions.
\citet{wu2025takes} argue that GRPO already belongs to a contrastive learning framework that DPO falls under.
However, this framework assumes an abstract signed-gradient template \citep{tao2022exploring} which is broadly permissive,
subsuming any method (including NCE) that pushes the weights in opposing directions for positive and negative outcomes.
In particular, it does not by itself specify any concrete objective, and their main algorithmic instantiation is GRPO with group size 2.
In contrast, we develop a new contrastive learning alglorithm.

Online DPO is a promising alternative to GRPO for RL post-training.
It aims to preserve the training stability of offline DPO while improving its performance by refreshing preference pairs from the evolving policy.
Like ReNCE, online DPO is explicitly contrastive.
However, empirical results are inconclusive.
\citet{su2025trust} argue for the necessity of various modifications to online DPO,
but their final results do not include end-to-end performance on standard math benchmarks (only post-tuning an R1-distilled checkpoint).
\citet{lanchantin2025bridging}, in contrast, argue that online/semi-online DPO is already competitive with GRPO,
but the reported performance of both is rather modest (e.g., $<60$ on Math500) because they train on in-house data, making strong conclusions difficult.
In this paper, we directly show that ReNCE outperforms online DPO.

There is prior work on extending offline DPO from pairwise preferences to multiclass ranking objectives via NCE-style losses.
For instance, \citet{chen2024noise} show that a reward-weighted softmax loss can outperform pairwise preference losses
on UltraFeedback, where each prompt is associated with a fixed set of candidate responses and scalar reward annotations.
In contrast, we consider the more challenging setting of on-policy post-training for reasoning, and present a method competitive with strong modern baselines.

\section{Method}

Given a prompt $x$, an LLM with parameters $\theta$ defines a policy $\pi_\theta(y|x)$ over reasoning trajectories $y$.
%In our context, $x$ is a mathematical problem and $y$ is the derivation of a solution;
The reward $r(x,y) \in [0, 1]$ is issued automatically based on the correctness and format of the solution.
At each iteration of ReNCE, we sample a group of $K$ iid candidates $\mathcal{G}(x) = \{ y_1 \ldots y_K \}$ from the previous policy $\pi_{\mathrm{old}}(\cdot|x)$,
partition them into highest-reward $\mathcal{P}(x)$ achieving $r_{\max}(x) = \max_{y' \in \mathcal{G}(x)} r(x,y')$ and the rest $\mathcal{N}(x)$,
and maximize the following per-prompt objective:
\begin{align}
  J(\theta) &=\frac{1}{|\mathcal{P}(x)|} \sum_{y \in \mathcal{P}(x)} \log \bigg( \frac{\exp(s_\theta(x,y))}{\sum_{y' \in \{y\} \cup \mathcal{N}(x)} \exp(s_\theta(x, y')) } \bigg) \notag \\
  %\max_\theta\; &\frac{1}{|\mathcal{P}(x)|} \sum_{y \in \mathcal{P}(x)} \log \bigg( \frac{e^{s_\theta(x,y)}}{\sum_{y' \in \{y\} \cup \mathcal{N}(x)} e^{s_\theta(x, y')} } \bigg) \notag\\
  &\hspace{25mm} - \lambda \mathrm{KL} (\pi_{\mathrm{old}}(\cdot|x) || \pi_{\theta}(\cdot|x) ) \label{eq:rence}
\end{align}
where the score $s_\theta(x, y) \in \R$ of a prompt-trajectory pair under the model is computed by the shifted log-ratio:
\begin{align}
  s_\theta(x, y) = \beta \log \frac{\pi_\theta(y|x)}{\pi_{\mathrm{old}}(y|x)} + \alpha m(x, y)
  \label{eq:score}
\end{align}
We use an adaptive margin $m(x, y) = r_{\max}(x) - r(x, y)$.
The scalars $\lambda, \beta, \alpha \in \R$ are hyperparameters.
The training procedure under ReNCE is given in Algorithm~\ref{alg:rence}.
In the remaining section, we flesh out the key pieces of this objective.

\begin{algorithm}[t!]
\caption{\thiswork{} Training Iteration}
\label{alg:rence}
\begin{algorithmic}[1]
\REQUIRE Train dataloader $\mathcal{D}$, group size $K$, batch size $B$, update steps $n_{\text{\tiny update}}$, filtering thresholds $t_{\mathrm{hard}},t_{\mathrm{easy}} \in [0, 1]$, mastery threshold $t_{\mathrm{master}}$
\ENSURE Updated policy $\pi_\theta$.
\STATE Maintain a prompt cache $\mathcal{Q}$.
\STATE Initialize the on-policy anchor $\pi_{\mathrm{old}} \gets \pi_\theta$.
\STATE Initialize a working prompt stack $\mathcal{S}\gets[\ ]$.
\WHILE{$|\mathcal{S}|<B$}
    \STATE $\tilde{\mathcal{B}} \gets \mathcal{Q}$ if $|\mathcal{Q}| > 0$; otherwise $\tilde{\mathcal{B}} \gets \mathcal{D}.\mathrm{batch}()$.
    \STATE $\mathcal{Q}\gets[\ ]$ \COMMENT{cache consumed if used}
    \FORALL{$x \in \tilde{\mathcal{B}}$}
        \STATE Sample $\mathcal{G}(x)\gets\{y_i\}_{i=1}^K \sim \pi_{\mathrm{old}}(\cdot\mid x)$ and compute rewards $\mathcal{R}(x) \gets \{r(x,y_i)\}_{i=1}^K$
        \STATE Compute $\rho(x) \gets |\mathcal{P}(x)|/|\mathcal{G}(x)|$
        \IF{$\rho(x)>t_{\mathrm{master}}$}
            \STATE Mark $x$ as mastered \COMMENT{permanently removed}
        \ELSIF{$t_{\mathrm{hard}}<\rho(x)\le t_{\mathrm{easy}}$}
            \STATE Append $(x,\mathcal{G}(x),\mathcal{R}(x))$ to $\mathcal{S}$
        \ENDIF
    \ENDFOR
\ENDWHILE
\STATE $\mathcal{B}\gets \mathcal{S}[:B]$
\STATE $\mathcal{Q} \gets \mathcal{S}[B:]$
\FOR{$j=1,2,\dots,n_{\text{\tiny update}}$}
    \STATE Update $\theta$ by minimizing (\ref{eq:rence}) on batch $\mathcal{B}$.
\ENDFOR
\end{algorithmic}
\end{algorithm}

\subsection{Trajectory Partitioning}

NCE assumes the specification of a ``correct'' trajectory in the candidate trajectories $\mathcal{G}(x)$.
Since our goal is to reinforce the best reasoning behavior, we treat all highest-reward trajectories as correct and define the dynamic positive set
\begin{align*}
  \mathcal{P}(x) = \{y \in \mathcal{G}(x):\; r(x,y) = r_{\max}(x)\}
\end{align*}
with an associated negative set $\mathcal{N}(x) = \mathcal{G}(x) \backslash \mathcal{P}(x)$.
This is a form of hard negative mining because the negative trajectories also evolve during training, becoming increasingly refined and harder to distinguish from positives.
Hard negative mining is shown to have empirical and theoretical benefits in the NCE literature \citep{zhang2021understanding,gillick-etal-2019-learning}.

Note that no meaningful learning is possible in the degenerate case $\mathcal{P}(x) = \mathcal{G}(x)$ (i.e., all candidates have the same reward).
But this is equally true for GRPO since all candidates have zero advantage.
We assume that it is always possible to find some prompts with nonzero reward variance at any training step, which motivates dynamic prompt filtering.

\subsection{Dynamic Prompt Filtering}
\label{sec:filtering}

\citet{yu2025dapo} show the importance of filtering prompts with zero reward variance (in short, ZV filtering) in GRPO to keep the effective batch size stable.
We generalize this mechanism by introducing tunable thresholds $t_{\mathrm{hard}}, t_{\mathrm{easy}} \in [0, 1]$
and including a prompt $x$ in the batch only if
\begin{align}
  t_{\mathrm{hard}} < \rho(x) \leq t_{\mathrm{easy}} \label{eq:pr-filtering}
\end{align}
where $\rho(x) = |\mathcal{P}(x)|/|\mathcal{G}(x)|$ is the ratio of positive trajectories.
ZV filtering is a special case with $t_{\mathrm{hard}} = 0$ and $t_{\mathrm{easy}}$ set close to 1.
We keep $t_{\mathrm{hard}} = 0$ but set $t_{\mathrm{easy}} \ll 1$ to focus learning on prompts that are not ``too easy'' for the current model.
In experiments, we find that ReNCE significantly benefits from setting $t_{\mathrm{easy}} = 0.5$ (i.e., only use prompts with at least 50\% incorrect rollouts) compared to naive ZV filtering (Table~\ref{tab:loss_filter}).

As in DAPO, prompt filtering is applied dynamically at each iteration.
We oversample prompts to form a candidate pool and retain only those satisfying \eqref{eq:pr-filtering}.
We repeatedly sample and filter prompts until the target training batch size is reached.
Note that prompts that fail the criterion at a given step may still be reconsidered in future iterations as the policy evolves,
inducing an implicit curriculum that excludes too-difficult problems in early training and too-easy problems in later stages.

In addition, we apply mastered-prompt filtering that permanently removes prompts with $\rho(x) > 0.8$ from the training pool to save rollout compute and focus learning on prompts that continue to provide informative signals.
This is inspired by prior work \citep{zheng2025act} observing that many easy (effectively mastered) prompts remain uninformative across training.
Unlike their stochastic backoff strategy, we use irreversible pruning and find it sufficient in our setting.

\subsection{Multi-Label NCE} \label{sec:mnce}

Conventional NCE assumes a single positive label while we have potentially multiple positive labels $\mathcal{P}(x) \subset \mathcal{G}(x)$ in our NCE formulation.
We adopt the multi-label NCE (mNCE) objective of \citet{zhang2022entqa} which treats each positive as independent and optimizes
\begin{align*}
  \max_\theta\; \sum_{y \in \mathcal{P}(x)} \log \bigg( \frac{\exp(s_\theta(x,y))}{\sum_{y' \in \{y\} \cup \mathcal{N}(x)} \exp(s_\theta(x, y')) } \bigg)
\end{align*}
Since the negative set $\mathcal{N}(x)$ is already disjoint from $\mathcal{P}(x)$, the mNCE objective does not require per-label negative set as in \citet{zhang2022entqa}
and is particularly simple, yielding $|\mathcal{P}(x)|$ independent instances of NCE per prompt.
We normalize the objective by $|\mathcal{P}(x)|$ in \eqref{eq:rence} to equally weight all positive trajectories within a group and prevent groups with many positives from dominating the training signal.

\subsection{Prompt-Trajectory Score}

The objective (\ref{eq:rence}) is usable with any backpropagation-friendly definition of prompt-trajectory score $s_\theta(x, y) \in \R$.
Here, we follow DPO which derives an implicit reward of a trajectory as the log-ratio between the current policy and a frozen reference policy (plus a log-partition term).
Since we do not use a reference policy, we use the previous policy $\pi_{\mathrm{old}}$ which is used to generate the candidates $\mathcal{G}(x)$, yielding the base score
\begin{align}
   s_{\theta}^{\mathrm{DPO}}(x, y) = \beta \log \frac{\pi_\theta(y|x)}{\pi_{\mathrm{old}}(y|x)} \label{eq:score-base}
\end{align}
which has also been adopted in recent work on contrastive learning for offline LLM alignment~\citep{chen2024noise}.
Here $\pi(y|x) = \prod_{t=1}^{|y|} \pi(y_t|x, y_{<t})$ and $\beta \in \R$ is a hyperparameter.
We omit the log-partition term used in DPO's implicit reward formulation, as softmax is invariant to translation. 

A key advantage of this log-ratio formulation is its inherent resistance to length bias.
Raw log-probabilities $\log \pi_\theta(y|x)$ tend to penalize longer sequences, while simple length averaging $\frac{1}{|y|} \log \pi_\theta(y|x)$ may over-correct.
By taking the log-ratio with $\pi_{\mathrm{old}}$, length-dependent effects shared by both policies largely cancel out during the subtraction in log-space,
allowing the score to emphasize relative reasoning quality rather than sequence length.

\subsubsection{Dynamic reward-scaled margin}

A potential limitation of (\ref{eq:score-base}) is its insensitivity to rewards,
which can be useful in fine-grained control of learning signals and tie breaking.
To this end, we introduce an adaptive margin
\begin{align}
  m(x, y) = r_{\max}(x) - r(x, y) \label{eq:margin}
\end{align}
yielding the final score in \eqref{eq:score} as
\begin{align*}
   s_\theta(x, y) = \beta \log \frac{\pi_\theta(y|x)}{\pi_{\mathrm{old}}(y|x)} + \alpha m(x, y)
\end{align*}
where $\alpha \in \R$ is a hyperparameter.
The margin (\ref{eq:margin}) measures the reward gap between the best trajectory in the group and the current trajectory.
Thus higher-quality negatives receive smaller margins, while lower-quality negatives receive larger margins and are pushed further from the positives.
Margin-based objectives are known to improve the generalization performance of classifiers~\citep{boser1992training,cortes1995support,turner2012bradley}.

Beyond enlarging decision boundaries, the margin also ameliorates gradient vanishing at early training stages. When $\pi_\theta = \pi_{\mathrm{old}}$, the DPO-style implicit reward (\ref{eq:score-base}) is zero for all trajectories,
yielding uniform softmax weights and weak updates. The added term $\alpha m(x, y)$ breaks this symmetry, inducing gradients that immediately emphasize higher-quality reasoning paths.

\subsection{Trust Region}\label{sec:kl}

Similar to PPO and TRPA \citep{schulman2017proximal,su2025trust}, ReNCE uses a KL penalty $\mathrm{KL} (\pi_{\mathrm{old}}(\cdot|x) || \pi_{\theta}(\cdot|x) )$ between the current policy $\pi_\theta$ and the old policy $\pi_{\mathrm{old}}$ to enforce a trust region.
In particular, we adopt the adaptive KL penalty scheme in \citet{ziegler2019fine} which has been shown to be more effective than using a fixed KL coefficient.
We estimate the KL divergence using the unbiased estimator proposed by~\citet{schulman2020kl}.

\section{Experiments}

\subsection{Setup}

\paragraph{Training Datasets.}
We construct our training set by combining the DAPO dataset~\citep{yu2025dapo} with the MATH dataset~\citep{hendrycks2measuring}.
This combination exposes the model to both integer-answer problems and more general symbolic, LaTeX-formatted mathematical solutions~\citep{zheng2025act}.

\paragraph{Evaluation.}
All models are evaluated on six widely used and challenging mathematical reasoning benchmarks: MATH500~\citep{hendrycks2measuring}, AIME24~\citep{aime24}, AIME25~\citep{aime25}, AMC~\citep{aimo_amc_hf_2024},
Minerva Math~\citep{lewkowycz2022solving}, and OlympiadBench~\citep{he2024olympiadbench}.
Following previous work~\citep{zheng2025act,wang2025reinforcement}, we evaluate all models using a sampling temperature of 0.7 and repeat each test set four times to improve evaluation stability,
reporting pass@1 (avg@4) based on the checkpoint with the best average validation performance.

\paragraph{Implementation Details.}
We implement \thiswork{} using the \textsc{verl} framework~\citep{sheng2024hybridflow} with vLLM~\citep{kwon2023efficient} for rollout generation. For each prompt, we sample $K{=}8$ rollouts, and use reward scores primarily for relative ranking within each rollout group rather than as an absolute optimization target. Following standard practice in RLVR~\citep{guo2025deepseek}, we assign the reward of prompt--rollout pair $(x,y)$ as
\begin{align*}
  r(x, y) = \begin{cases}
  1 &\text{if $\mathrm{Boxed}(y) \wedge \mathrm{Correct}(x,y)$} \\
  0.1 &\text{if $\mathrm{Boxed}(y) \wedge \neg \mathrm{Correct}(x,y)$} \\
  0 &\text{otherwise,}
  \end{cases}
\end{align*}
where $\mathrm{Boxed}(y)$ indicates that $y$ wraps the final answer in \texttt{\textbackslash boxed\{\}}, and $\mathrm{Correct}(x,y)$ checks whether the boxed answer is correct for problem $x$. We set $\beta{=}0.1$ and $\alpha{=}0.5$ in Eq.~\ref{eq:score}. We use the maximum prompt length of 1536 tokens and the maximum response length of 6656 tokens, for a total context length of 8192 tokens. We use a unified prompt template for both training and evaluation~\citep{zheng2025act}:
\begin{center}
    \fbox{
        \parbox{7.5cm}{ % Adjust width (e.g., 8cm) as needed
            %\centering % Center the text within the parbox
            \texttt{\small Please solve the following math problem: \{problem\}. The assistant first thinks about the reasoning process step by step and then provides the user with the answer. Return the final answer in \textbackslash boxed\{\} tags, for example \textbackslash boxed\{1\}. Let's solve this step by step.}
        }
    }
\end{center}
For optimization, we use a training batch size of 512 prompts per iteration and a mini-batch size of 32 prompts per policy update. Concretely, we process 2 prompts per GPU on 8 GPUs and use gradient accumulation with 2 steps to reach the 32-prompt mini-batch. We optimize the actor with AdamW~\citep{loshchilovdecoupled} using a constant learning rate of $10^{-6}$, $\beta_1{=}0.9$, $\beta_2{=}0.999$, and weight decay $0.01$, with gradient clipping at max norm 1.0. Model parameters are wrapped with FSDP~\citep{zhao2023pytorch} for distributed training; we enable gradient checkpointing and use bfloat16 precision throughout. For the adaptive KL coefficient (Section~\ref{sec:kl}), we use an initial KL coefficient of 0.001, a target KL of 0.01, and a horizon of 25600. All experiments are initialized from the Qwen3-4B pretrained model and run on 8$\times$A100 (80GB) GPUs for 6 days. We focus on Qwen3-4B due to compute constraints and leave larger models (e.g., Qwen3-8B) to future work.

\begin{table*}[t!]
\caption{Main results on six mathematical reasoning benchmarks. We report pass@1 (avg@4) in \% as mean (std) across 4 repeats; both mean and std are in \%. Best results are in \textbf{bold} and second best are \uline{underlined}.}
\label{tab:main_results_mean_std_paren}
\centering
\begin{adjustbox}{width=\textwidth}
\begin{tabular}{l|cccccc|c}
\toprule
Method & Olympiad & Minerva & AIME25 & AIME24 & MATH-500 & AMC & Avg \\
\midrule
No training
& 44.7 (7.0) & 41.7 (5.3) & 19.2 (6.0) & 32.5 (13.4) & 84.5 (5.5) & 55.7 (10.1) & 46.4 (7.9) \\
GRPO
& 58.5 (7.2) & 44.6 (5.1) & \textbf{48.3} (9.1) & 44.2 (18.9) & 92.5 (4.0) & 73.8 (12.0) & 60.3 (9.4) \\
DAPO
& \textbf{63.2} (6.8) & \uline{46.3} (6.2) & 35.8 (7.7) & \uline{49.2} (16.8) & \textbf{93.5} (4.0) & \textbf{82.5} (7.6) & \uline{61.8} (8.2) \\
Semi-online DPO
& 59.6 (5.7) & \textbf{46.6} (5.7) & 31.7 (9.1) & 37.5 (8.9) & \textbf{93.5} (2.8) & 78.0 (7.6) & 57.8 (6.6) \\
\thiswork{}
& \uline{62.6} (9.7) & \textbf{46.6} (7.0) & \uline{43.3} (9.1) & \textbf{53.3} (17.0) & \uline{93.2} (4.2) & \uline{78.9} (10.8) & \textbf{63.0} (9.6) \\
\bottomrule
\end{tabular}
\end{adjustbox}
\label{tab:main_results}
%\vskip -0.1in
\end{table*}

% \begin{table*}[t!]
% \caption{Main results on six mathematical reasoning benchmarks. We report pass@1 (avg@4) in \%. Best results are in \textbf{bold} and second best are \uline{underlined}.}
% \centering
% \begin{adjustbox}{width=\textwidth}
% \begin{tabular}{l|cccccc|c}
% \toprule
% Method & Olympiad & Minerva & AIME25 & AIME24 & MATH-500 & AMC & Avg \\
% \midrule
% No training        & 44.7 & 41.7 & 19.2 & 32.5 & 84.5 & 55.7 & 46.4 \\
% GRPO               & 58.5 & 44.6 & \textbf{48.3} & 44.2 & 92.5 & 73.8 & 60.3 \\
% DAPO               & \textbf{63.2} & 46.3 & 35.8 & \uline{49.2} & \textbf{93.5} & \textbf{82.5} & \uline{61.8} \\
% Semi-online DPO    & 59.6 & \textbf{46.6} & 31.7 & 37.5 & \textbf{93.5} & 78.0 & 57.8 \\
% \thiswork{}        & \uline{62.6} & \textbf{46.6} & \uline{43.3} & \textbf{53.3} & 93.2 & \uline{78.9} & \textbf{63.0} \\
% \bottomrule
% \end{tabular}
% \end{adjustbox}

% \label{tab:main_results}
% %\vskip -0.1in
% \end{table*}

\paragraph{Baselines.}

% initial Qwen3-4B, grpo, dapo and semi Online DPO as the baselines
We compare against the following four baselines:
\begin{itemize}
    \item \textbf{No training.} The Qwen3-4B pretrained model without any further training.
    \item \textbf{GRPO}~\citep{shao2024deepseekmath}. Qwen3-4B trained on our dataset using the GRPO algorithm.
    \item \textbf{DAPO}~\citep{yu2025dapo}. A state-of-the-art extension of GRPO that incorporates several effective techniques,
      including higher clipping thresholds, dynamic sampling to remove zero-variance samples, token-level policy gradient loss, and overlong response penalties.
    \item \textbf{Semi-online DPO}~\citep{lanchantin2025bridging}. A DPO variant that generates rollouts online and updates the model using the DPO loss for $s$ steps per rollout batch, equivalent to synchronizing the reference model every $s$ steps. \citet{lanchantin2025bridging} show that the semi-online setting ($s>1$) matches or outperforms fully online DPO ($s=1$), thus we focus on the former. Similar to DAPO, we apply dynamic sampling to filter zero-variance samples to ensure that at least one positive--negative pair can be formed. We use the Group-DPO objective that averages the DPO loss over all positive--negative pairs within each rollout group ($K=8$); empirically, it performs comparably to using a single sampled pair while providing a deterministic way to exploit all in-group comparisons.

\end{itemize}

For \thiswork{}, DAPO, and semi-online DPO, we perform 16 optimization steps per rollout batch for fair comparison. For GRPO, we find that fewer update steps ($\leq 4$) yield substantially better performance than 16, and therefore use 4 update steps per rollout batch. Across all methods, we use the same prompt batch size of 512 prompts per rollout iteration. We train GRPO for 200 rollout iterations, DAPO and semi-online DPO for 80 iterations, and \thiswork{} for 50 iterations, reflecting the increasingly aggressive filtering strategies used by the latter methods.
% also add some details about num update steps per training step

\subsection{Results}

Table~\ref{tab:main_results} summarizes the main results. All training methods substantially improve over the pretrained model, demonstrating the effectiveness of post-training on mathematical reasoning tasks. \thiswork{} achieves the highest average performance across the six benchmarks and maintains consistently strong results on each task, ranking best or near-best throughout. GRPO and DAPO also perform strongly and significantly outperform the online DPO baseline, aligning with prior findings that reward-based policy gradient methods are generally more effective than preference-based DPO approaches on reasoning-intensive problems~\citep{zhangonline,su2025trust}. Moreover, DAPO consistently outperforms GRPO, validating the effectiveness of its additional refinements built on top of GRPO. We report mean (std) over four repeated evaluations (avg@4); while variance is non-negligible on the hardest benchmarks (e.g., AIME24), the overall trend is consistent, with average-score std in the 6.6--9.6 point range across methods. Compared with these strong GRPO-style baselines, \thiswork{} attains the best average score across the benchmarks, demonstrating the effectiveness of our explicit contrastive learning approach for LLM reasoning.

% Table~\ref{tab:main_results} summarizes the main results. All training methods substantially improve over the pretrained model, demonstrating the effectiveness of post-training on mathematical reasoning tasks. \thiswork{} achieves the highest average performance across the six benchmarks and maintains consistently strong results on each task, ranking best or near-best throughout. GRPO and DAPO also perform strongly and significantly outperform the online DPO baseline, aligning with prior findings that reward-based policy gradient methods are generally more effective than preference-based DPO approaches on reasoning-intensive problems~\citep{zhangonline,su2025trust}. Moreover, DAPO consistently outperforms GRPO, validating the effectiveness of its additional refinements built on top of GRPO. Compared with these strong GRPO-style baselines, \thiswork{} delivers stronger overall performance and attains the best average score across the benchmarks. Overall, these results demonstrate the effectiveness of our explicit contrastive learning approach for LLM reasoning.

\section{Analysis}

\begin{table*}[t]
\caption{Component ablations of \thiswork{}. We report pass@1 (avg@4) in \%. Best results are in \textbf{bold}.}
\centering
\begin{adjustbox}{width=\textwidth}
\begin{tabular}{l|cccccc|c}
\toprule
Method & Olympiad & Minerva & AIME25 & AIME24 & MATH-500 & AMC & Avg \\
\midrule
\thiswork{}
& 62.6 & \textbf{46.6} & 43.3 & \textbf{53.3} & 93.2 & 78.9 & \textbf{63.0} \\

No Trust-Region KL
& \textbf{62.8} & 44.9 & \textbf{44.2} & 45.8 & 93.2 & 77.7 & 61.4 \\

No Margin
& 62.6 & 46.0 & \textbf{44.2} & 51.7 & \textbf{93.4} & 79.2 & 62.9 \\

Constant Margin
& 60.0 & 46.1 & 37.5 & 52.5 & 93.1 & 78.0 & 61.2 \\

Iterative
& 59.7 & 45.7 & 41.7 & 42.5 & 93.1 & \textbf{79.8} & 60.4 \\
\bottomrule
\end{tabular}
\end{adjustbox}

\label{tab:ablation}
\vskip -0.1in
\end{table*}

\begin{table*}[t]
\caption{Effect of the positive-ratio upper bound $t_{\mathrm{easy}}$ (holding $t_{\mathrm{hard}} = 0$ fixed). We report pass@1 (avg@4) in \%. Best results are in \textbf{bold}.}
\centering
\begin{adjustbox}{width=\textwidth}
\begin{tabular}{l|cccccc|c}
\toprule
Method & Olympiad & Minerva & AIME25 & AIME24 & MATH-500 & AMC & Avg \\
\midrule
\thiswork{} ($t_{\mathrm{easy}}{=}0.5$)
& 62.6 & 46.6 & \textbf{43.3} & \textbf{53.3} & 93.2 & 78.9 & \textbf{63.0} \\

\thiswork{} ($t_{\mathrm{easy}}{=}0.99$, ZV)
& 59.6 & \textbf{47.1} & 35.0 & 45.0 & 92.5 & 74.4 & 58.9 \\

\midrule
DAPO ($t_{\mathrm{easy}}{=}0.5$)
& 62.3 & 46.0 & \textbf{43.3} & 44.2 & \textbf{93.6} & 79.5 & 61.5 \\

DAPO ($t_{\mathrm{easy}}{=}0.99$, ZV)
& \textbf{63.2} & 46.3 & 35.8 & 49.2 & 93.5 & \textbf{82.5} & 61.8 \\
\bottomrule
\end{tabular}
\end{adjustbox}

\label{tab:loss_filter}
\vskip -0.1in
\end{table*}

\begin{table*}[t]
\centering
\caption{Comparison of different contrastive objectives. We report pass@1 (avg@4) in \%. Best results are in \textbf{bold}.}
\label{tab:objective}
\begin{adjustbox}{width=\textwidth}
\begin{tabular}{l|cccccc|c}
\toprule
Objective & Olympiad & Minerva & AIME25 & AIME24 & MATH-500 & AMC & Avg \\
\midrule
mNCE
& \textbf{62.6} & 46.0 & \textbf{44.2} & \textbf{51.7} & \textbf{93.4} & \textbf{79.2} & \textbf{62.9} \\

Softmax 
& 60.0 & 45.9 & 37.5 & 48.3 & 92.9 & \textbf{79.2} & 60.6 \\

Random pair pairwise
& 60.0 & \textbf{46.8} & 35.8 & 48.3 & 93.2 & 77.1 & 60.2 \\

All-pairs pairwise
& 60.5 & 46.0 & 37.5 & 46.7 & 92.4 & 78.0 & 60.2 \\

\bottomrule
\end{tabular}
\end{adjustbox}
\vskip -0.1in
\end{table*}

\subsection{Component Ablations}\label{sec:compo}

We study the contribution of key components in \thiswork{}. Table~\ref{tab:ablation} reports the results, where we vary one component at a time while keeping all others identical to the \thiswork{} baseline. All methods are initialized from the Qwen3-4B pretrained model and trained for the same number of steps. We evaluate on the test benchmarks every 5 steps and report pass@1 (avg@4) using the checkpoint with the best average validation performance for each method. For fair comparison, we apply the same positive-ratio based filtering strategy to all variants.

Removing the trust-region KL penalty leads to a noticeable drop in average performance, with the largest degradation on AIME24. We further evaluate an \emph{iterative} variant that increases the prompt batch size and performs 128 policy updates between rollout generations while also removing the trust-region KL term. This setting is analogous to semi-online DPO~\citep{lanchantin2025bridging} with a larger synchronization interval ($s = 128$), and is substantially less on-policy than our default training ($s = 16$). Despite keeping other components unchanged (e.g., positive-ratio filtering) and matching the total number of policy update steps, this iterative variant performs significantly worse than both \thiswork{} and the no-KL ablation, indicating that staying near on-policy is critical.

For the margin component, we compare removing the margin term in Eq.~\ref{eq:rence} and replacing the reward-scaled margin with a constant margin (set to 1 with scaling $\alpha$). Interestingly, removing the margin yields performance close to the full model, whereas using a constant margin consistently hurts performance. We nevertheless retain the reward-scaled margin in \thiswork{} due to its lower variance across checkpoints and its ability to modulate contrastive strength and differentiate negatives of varying quality.

\subsection{Positive-Ratio Filtering}

We evaluate the impact of the positive-ratio based filtering strategy introduced in Section~\ref{sec:filtering}. We compare the widely used all-zero variance filtering heuristic from DAPO~\citep{yu2025dapo}, implemented by setting $t_{\mathrm{easy}}= 0.99$, against a stricter upper bound $t_{\mathrm{easy}}=0.5$, for both \thiswork{} and DAPO. We do not heavily tune $t_{\mathrm{easy}}$ in this work, as our goal is to isolate the qualitative effect of filtering rather than optimize hyperparameters.

Table~\ref{tab:loss_filter} shows that restricting the positive ratio to avoid overly easy prompts is crucial for \thiswork{}: using $t_{\mathrm{easy}}=0.5$ yields substantially better performance than the all-zero variance setting, whereas DAPO is comparatively insensitive to this choice. We hypothesize that this difference stems from the learning signal: \thiswork{} relies on explicit contrast between positive and negative trajectories, and prompts with very high positive ratios provide weak contrastive signal while potentially introducing noise. In contrast, DAPO optimizes an implicit contrastive objective through advantage normalization, making it less sensitive to the presence of easy prompts. Consistent with this interpretation, when using all-zero variance filtering for \thiswork{}, we observe initial gains during early training followed by divergence and degraded performance; A similar instability pattern also appears in our semi-online DPO experiments, where we likewise apply all-zero variance filtering to ensure valid positive--negative pairs and optimize an explicit contrastive objective by averaging over all positive--negative pairs within each rollout group.

\subsection{Contrastive Objective Variants}

We run experiments to compare different contrastive objectives, with results shown in Table~\ref{tab:objective}.
For controlled comparison, we fix the same positive-ratio filtering setting ($t_{\mathrm{easy}}=0.5$) across all methods and remove the margin term for simplicity,
isolating the effect of the contrastive loss itself.

Specifically, mNCE corresponds to the multi-label NCE objective in Section~\ref{sec:mnce}. Random-pair pairwise samples one positive trajectory from  $\mathcal{P}(x)$ and one negative trajectory from $\mathcal{N}(x)$ and applies a standard pairwise objective, analogous to vanilla DPO. All-pairs pairwise averages the pairwise loss over all positive--negative pairs within the rollout group, matching the Group-DPO objective in~\citet{lanchantin2025bridging}.
We also consider a softmax contrastive loss that, unlike our mNCE formulation, includes other positives in the normalization term---a common design choice in multi-label contrastive learning~\citep{khosla2020supervised}.

Overall, mNCE achieves the best performance, substantially outperforming the pairwise objectives. Interestingly, random-pair and all-pairs pairwise perform nearly identically in our setting, suggesting that enumerating all positive--negative pairs offers limited benefit beyond sampling a single pair for the pairwise loss. In contrast, mNCE provides a richer and more stable training signal by contrasting each positive against the full set of in-group negatives within a single normalized objective, which likely improves gradient quality and reduces sensitivity to pair selection. Moreover, compared with all-pairs pairwise, mNCE avoids quadratic pair enumeration by aggregating all negatives in one shot.
Finally, excluding other positives from the normalization term is important for performance, consistent with prior observations in~\citet{zhang2022entqa}.

% \subsection{Number of Rollouts}

% \begin{table*}[t]
% \caption{Effect of the number of rollouts $G$ per prompt. We report pass@1 (avg@4) in \%.}
% \centering
% \begin{adjustbox}{width=\textwidth}
% \begin{tabular}{lccccccc}
% \toprule
% Model & Olympiad & Minerva & AIME25 & AIME24 & MATH-500 & AMC & Avg \\
% \midrule
% \thiswork{}-8
% & 62.6 & 46.6 & 43.3 & 53.3 & 93.2 & 78.9 & \textbf{63.0} \\

% \thiswork{}-16 (running)
% & \multicolumn{7}{c}{running} \\

% \midrule
% DAPO-8
% & 63.2 & 46.3 & 35.8 & 49.2 & 93.5 & 82.5 & 61.8 \\

% DAPO-16 (running)
% & \multicolumn{7}{c}{running} \\
% \bottomrule
% \end{tabular}
% \end{adjustbox}

% \label{tab:rollouts}
% \vskip -0.1in
% \end{table*}

\section{Conclusions}
We present \thiswork{}, an explicit online contrastive learning approach for LLM reasoning. Unlike policy-gradient methods such as GRPO, which distinguish good and bad responses implicitly through advantage estimates over a group of rollouts, \thiswork{} directly contrasts positive and negative responses within each group using noise contrastive estimation. Across six challenging mathematical reasoning benchmarks, \thiswork{} achieves strong performance and outperforms competitive baselines including DAPO and online DPO.

%\section{Limitations}
%We discuss potential limitations of our setup. 
%Due to our compute constraints, %we are only able to evaluate up to 4-billion parameter models, 
%we use a relatively small number of training steps based on observed performance convergence. 
%We also do not heavily tune ReNCE hyperparameters (e.g., $t_{\mathrm{easy}}$).
%We demonstrate that ReNCE is already competitive under these limitations,
%and leave more exhaustive search over experimental setups beyond our constraints as future work.

%\section*{Acknowledgements}

\section*{Impact Statement}
This paper presents work whose goal is to advance the field
of Machine Learning, specifically reasoning capabilities of LLMs. 
There are many potential societal consequences of our work, 
none which we feel must be specifically highlighted here.


\bibliography{mybib}
\bibliographystyle{icml2026}

%%%%%%%%%%%%%%%%%%%%%%%%%%%%%%%%%%%%%%%%%%%%%%%%%%%%%%%%%%%%%%%%%%%%%%%%%%%%%%%
%%%%%%%%%%%%%%%%%%%%%%%%%%%%%%%%%%%%%%%%%%%%%%%%%%%%%%%%%%%%%%%%%%%%%%%%%%%%%%%
% APPENDIX
%%%%%%%%%%%%%%%%%%%%%%%%%%%%%%%%%%%%%%%%%%%%%%%%%%%%%%%%%%%%%%%%%%%%%%%%%%%%%%%
%%%%%%%%%%%%%%%%%%%%%%%%%%%%%%%%%%%%%%%%%%%%%%%%%%%%%%%%%%%%%%%%%%%%%%%%%%%%%%%
\newpage
\appendix
\onecolumn

% \section{More Experimental Details}\label{app:hyper}

% todo: add the algorithm here
%%%%%%%%%%%%%%%%%%%%%%%%%%%%%%%%%%%%%%%%%%%%%%%%%%%%%%%%%%%%%%%%%%%%%%%%%%%%%%%
%%%%%%%%%%%%%%%%%%%%%%%%%%%%%%%%%%%%%%%%%%%%%%%%%%%%%%%%%%%%%%%%%%%%%%%%%%%%%%%

\end{document}